\documentclass[11pt,a4paper]{article}
\usepackage[hyperref]{acl2020}
\usepackage{amsmath}
\usepackage{array}
\usepackage{arydshln}
\usepackage{diagbox}
\usepackage{times}
\usepackage{latexsym}
\usepackage{color}
\usepackage{makecell}
\usepackage{multirow}
\usepackage{listings}
\usepackage{soul}
\usepackage{url}
\usepackage{fixltx2e}
\usepackage{graphicx}
\usepackage{footnote}
\usepackage{gensymb}
\usepackage{microtype}

\aclfinalcopy % Uncomment this line for the final submission
\title{MultiReQA: A Cross-Domain Evaluation for \\Retrieval Question Answering Models}

\author{Mandy Guo\textsuperscript{$a$}\thanks{\hspace{2mm}Corresponding authors: \{xyguo, yinfeiy\}@google.com},~
Yinfei Yang\textsuperscript{$a$}\footnotemark[1],~
Daniel Cer\textsuperscript{$a$},~
Qinlan Shen\textsuperscript{$b$}\thanks{\hspace{2mm}Work done during an internship at Google Research.},~
and Noah Constant\textsuperscript{$a$} 
 \AND
  {\rm\textsuperscript{$a$}Google Research}\\Mountain View, CA, USA \And
  {\rm\textsuperscript{$b$}Carnegie Mellon University}\\Pittsburgh, PA, USA
}

\date{}

\begin{document}
\maketitle

\begin{abstract}
Retrieval question answering~(ReQA) is the task of retrieving a sentence-level answer to a question from an open corpus~\cite{reqa}.
This paper presents MultiReQA, a new multi-domain ReQA evaluation suite composed of eight retrieval QA tasks drawn from publicly available QA datasets\footnote{We released the sentence boundary annotation of MultiReQA: \url{https://github.com/google-research-datasets/MultiReQA}}. We provide the first systematic retrieval based evaluation over these datasets using two supervised neural models, based on fine-tuning BERT and USE-QA models respectively, as well as a surprisingly strong information retrieval baseline, BM25.
 % , by treating the sentence containing the ground-truth span as the sentence-level answer.
Five of these tasks contain both training and test data, while three contain test data only.
%Analysis 
Performance on the five tasks with training data shows that 
while a general model covering all domains is achievable, 
the best performance is often obtained by training exclusively on in-domain data.
% We will release the code for converting the retrieval dataset and evaluation .
\end{abstract}

% =====================================================================

\section{Introduction}

%In traditional machine reading for question answering or ``reading comprehension'', models are tasked with answering a question posed relative to a specific passage of text.
%However, 
Retrieval-based question answering (QA) investigates the problem of finding answers to questions from an open corpus~\cite{surdeanu-etal-2008-learning,yang-etal-2015-wikiqa,chen-etal-2017-reading,lee-etal-2019-latent,reqa,Chang2020Pre-training,ma2020zeroshot}.
There is a growing interest in building scalable end-to-end question answering systems for large scale retrieval.
Retrieval question answering~(ReQA)~\cite{reqa}, illustrated in Table \ref{tab:dataset_examples_w_context}, defines the task as \textit{directly} retrieving an answer sentence from a corpus.\footnote{This can be contrasted to a two stage approach that first retrieves supporting text and then identifies the correct answer span~\cite{chen-etal-2017-reading,lee-etal-2019-latent}}
Motivated by real applications such as Google’s Talk to Books~\footnote{\url{https://books.google.com/talktobooks/}}, where sentence-level answers from books are retrieved to answer users' queries, ReQA is different from traditional machine reading for question answering or ``reading comprehension'' which aims to extract a short answer span from a given passage. Rather than just identifying answers within a short preselected passage that is provided to the model effectively by an oracle, retrieving sentence-level answers from a large pool of candidates directly addresses the real-world problem of searching for answers within a corpus. Sentences retrieved as answers in this manner can be used directly to answer questions. Alternatively, retrieved sentences, as well as possibly the passages that contains them, can be provided to a traditional Open Domain QA model \cite{chen-etal-2017-reading,karpukhin2020dense}.

\begin{table}[t!]
\small
\centering
\begin{tabular}{| p{7cm} |}
\hline
    \textbf{Question:} In what year did Cortes send the first cochineal to Spain? \\
    \textbf{Answer in Context:} [...]
    It worked particularly well on silk, satin and other luxury textiles. \textbf{In 1523 Cortes sent the first shipment to Spain.} Soon cochineal began to arrive in European ports aboard convoys of Spanish galleons. \\
\hline
\end{tabular}
\caption{ReQA example drawn from SQuAD. The goal is to retrieve the answer sentence (\textbf{bolded}) from an open corpus based on the meaning of the sentence and the surrounding context.}
\label{tab:dataset_examples_w_context}
\end{table}

% =====================================================================

\begin{table*}[tp!]
    \small
    \centering
    \begin{tabular}{ p{1.5cm} | p{5.7cm} | p{7.5cm} } \hline
    \textbf{Dataset} & \textbf{Question} & \textbf{Answer}\\ \hline  \rule{-5pt}{2pt}
    SearchQA & At age 33 in 1804, he started a new symphony, his 5th, with a Da-Da-Da-Duhg & This is the first movement of Beethoven's 5th symphony.\\
    \rule{-5pt}{14pt}
    TriviaQA & From the Greek for color, what element, with an atomic number of 24, uses the symbol Cr? & Rubies and emeralds also owe their colors to chromium compounds. \\ \rule{-5pt}{14pt}
    HotpotQA & Lenny Young is a collaborator on the stop motion film released in what year? & Chicken Run is a 2000 stop-motion animated comedy film produced by the British studio Aardman Animations.  \\
    \rule{-5pt}{14pt}
    NQ & when was the last episode of vampire diaries aired & The series ran from September 10, 2009 to March 10, 2017 on The CW. \\ \rule{-5pt}{14pt}
    SQuAD & what decade did house music hit the mainstream in the us? & The early 1990s additionally saw the rise in mainstream US popularity for house music. \\\hline  \rule{-5pt}{2pt}
    BioASQ & What chromosome is affected in Turner's syndrome? & The origin of sSMC of Turner syndrome with 45, X/46, X, + mar karyotype was almost all from sex chromosomes, and rarely from autosomes. \\ \rule{-5pt}{14pt}
    % DuoRC & What game does Bugs suggest to Sam? &  This leads Bugs to make the odd suggestion to Sam to play a game of Russian Roulette and hand a gun to Sam. \\ \rule{-5pt}{14pt}
    R.E. & Which year is Bird Girl and the Man Who Followed the Sun released? & Bird Girl and the Man Who Followed the Sun is a 1996 novel by Velma Wallis. \\ \rule{-5pt}{14pt}
    TextbookQA & which nervous system disease causes seizures? & Epilepsy is a disease that causes seizures. \\ \hline
    \end{tabular}
    \caption{Example questions and answers from each dataset. }
    \label{tab:dataset_examples}
\end{table*}

We introduce a new common evaluation suite and strong baselines for ReQA across eight publicly available QA tasks.
Five \emph{in-domain} tasks include training and test data, while three \emph{out-of-domain} tasks contain only test data.
%In-domain tasks include training and test data, while out-of-domain tasks only contain test data.
Our experiments investigate using two competitive neural models, based on BERT~\cite{devlin-etal-2019-bert} and USE-QA~\cite{yang2019multilingual}, respectively, and BM25, a strong information retrieval baseline.
%The context of candidate answer sentences is considered during the evaluation, e.g.~the answer sentence is concatenated with its surrounding context when using the BM25 model.
BM25 performs surprisingly well on many retrieval question answering tasks, achieving the best performance on two of five in-domain tasks and all three out-of-domain tasks.
%\footnote{Text pre-processing details affect BM25 performance considerably; we reach the best performance using casing normalization and WordPiece tokenization, which has not been combined with BM25 in previous work to our knowledge.}
%The neural models are trained on the training set of each in-domain task.
Neural models achieve the highest performance on three of five in-domain tasks, outperforming BM25 by a wide margin on tasks with less token overlap between question and answer.
%, for example improving precision at 1 (P@1) by +11.68 and +12.68 on HotpotQA and Natural Questions respectively.
%The results indicate that training on in-domain data is important for the neural models.
%Experiments also show that the BM25 approach relies more on the answer sentence context.
%When removing the context, the average P@1 performance of BM25 drops 4.45, which is much larger than the performance drop of the best neural model (2.42).
Comparing general models trained on a mixture of QA training sets to specialized in-domain models trained on a single QA task reveals that models trained jointly on multiple datasets rarely outperform those trained on only in-domain data.

\section{Retrieval QA (ReQA)}
\label{sec:reqaoverview}

ReQA formalizes the retrieval-based QA task as the identification of a sentence in-context that answers a provided question~\cite{reqa}. 
Retrieval QA models are evaluated using Precision at 1 (P@1) and Mean Reciprocal Rank (MRR).
The P@1 score tests whether the true answer sentence appears as the top-ranked candidate\footnote{Retrieval models are often measured by P@N (N=1,3,5,10). However, as our main concern is whether the question is correctly answered, we focus on P@1.}.
MRR, introduced for the evaluation of retrieval based QA systems \cite{Voorhees:2001:TQA:973890.973895,radev-etal-2002-evaluating}, is calculated as $\textrm{MRR} = \frac{1}{N} \sum_{i=1}^{N} \frac{1}{\textit{rank}_i}$, where $N$ is the total number of questions, and \emph{rank}$_i$ is the rank of the first correct answer for the $i$th question. 
% Both P@1 and MRR are shown as percentages in the tables.
% =====================================================================

\section{Multi-domain  ReQA (MultiReQA)}
\label{sec:dataset}

The multi-domain ReQA (MultiReQA) test suite is composed of select datasets drawn from the MRQA shared task \cite{fisch-etal-2019-mrqa}.\footnote{We exclude NewsQA, RACE, DROP, and DuoRC, as the majority of their questions are underspecified when taken out of their original context, making them inappropriate for a large-scale retrieval evaluations.} We follow the training, in-domain test, out-of-domain test splits defined in MRQA. The individual datasets are described below:

\vspace{-1mm}
\paragraph{SearchQA} Jeopardy question-answer pairs augmented with text snippets retrieved by Google \cite{searchqa}.

\vspace{-1mm}
\paragraph{TriviaQA} Trivia enthusiasts authored question-answer pairs.
% Question writers know the answers to the questions.
Answers are drawn from Wikipedia and Bing web search results, excluding trivia websites \cite{triviaqa}.

\vspace{-1mm}
\paragraph{HotpotQA} Wikipedia question-answer pairs.
% Question writers know the answers and supporting text.
This dataset differs from the others in that the questions require reasoning over multiple supporting documents \cite{hotpotqa}.

\vspace{-1mm}
\paragraph{SQuAD 1.1} Wikipedia question-answer pairs 
% Question writers know the answers and supporting text 
\cite{squad}.

\vspace{-1mm}
\paragraph{NaturalQuestions (NQ)} Questions are real queries issued by multiple users to Google search that retrieve a Wikipedia page  in the top five search results. Answer text is drawn from the search results \cite{naturalquestions}.

\vspace{-1mm}
\paragraph{BioASQ} Bio-medical question-answer pairs with answers annotated by domain experts and drawn from research articles \cite{bioasq}.

\vspace{-1mm}
\paragraph{RelationExtraction (R.E.)} Entity relation question-answer pairs, created by slot filling using the WikiReading dataset \cite{reqa}.

\vspace{-1mm}
\paragraph{TextbookQA} Multi-modal question-answer pairs taken from middle school science curricula \cite{textbookqa}.

% Our retrieval task format consists of questions, and their answer-context pairs as candidates. 
Table \ref{tab:dataset_examples} provides example question-answer sentence pairs. %\footnote{Examples from other datasets are included in Appendix.}
Datasets are converted from a span identification task to sentence-level retrieval. The questions from the original data are used without modification. Supporting documents are split into sentences using NLTK. All resulting sentences become retrieval candidates.
Answer spans are used to identify the sentences containing the correct answers.
\footnote{Retrieval candidates other than the sentence identified by the answer span could also provide the correct answer to a question. We investigate the prevalence of such false negatives in our subsequent analysis (6.4). As the datasets SearchQA, TriviaQA and HotpotQA contain special tags [DOC], [PAR], [SEP], and [TLE], we perform dataset-specific pre-processing to handle context splitting and tag removal.
%\footnote{The details of pre-processing individual datasets are in Appendix A.}.
TriviaQA has [DOC] [TLE] [PAR] tags, but with no clear divisions to mark where the span of each kind of tags ends. We remove all the tags, and tokenize the article as if it does not have special tags. SearchQA uses [DOC] to separate the supporting snippets, [TLE] to mark the start of title, and [PAR] to mark start of the snippet content. We treat contents between two [DOC] tags as individual context. We then use NLTK to split the sentences within each context. The contents between [TLE] and [PAR] are used as a title feature. If the answer appears in the title feature, we do not add it as a positive answer. There are about 500 examples where the answer span is only in the title span, and we remove the corresponding questions. We follow the same procedure for HotpotQA, which uses [PAR] to separate supporting documents, and [SEP] to separate title and document content.} 
Spans covering multiple sentence are excluded.\footnote{This is typically due to sentence splitting errors by NLTK.} 
% \paragraph{Limitations of sentence tokenization tool.}
Tables \ref{tab:dataset_stats} and \ref{tab:dataset_analysis} provide dataset statistics.

\begin{table}[t!]
    \small
    \centering
    \begin{tabular}{ p{1.45cm} | r | r r r  } \hline
    \multirow{3}{*}{\textbf{Dataset}} & \multirow{3}{*}{\textbf{Train}} & \multicolumn{3}{c}{\textbf{Test}}  \\ 
    & & Ques. & Cand. & \makecell{Avg. ans. \\ per ques.} \\ \hline
    SearchQA      & 629,160 & 16,476 & 454,836  & 5.47  \\
    TriviaQA      & 335,659 & 7,776 & 238,339 & 5.46  \\
    HotpotQA      & 104,973 & 5,859 & 52,191 & 1.69  \\
    SQuAD         & 87,133 & 10,485 & 10,642 & 1.09  \\
    NQ            & 106,521 & 4,131 & 22,118 & 1.06 \\
    BioASQ        & - & 1,503 & 14,158 & 2.85  \\
    R.E.          & - & 2,945 & 3,301 & 1.00 \\
    TextbookQA    & - & 1,497 & 3,701 & 3.31 \\
    \hline
    \end{tabular}
    \caption{Statistics for each constructed dataset: \# of training pairs, \# of questions, \# of candidates, and average \# of answers per question.}
    \label{tab:dataset_stats}
\end{table}

\begin{table}[t!]
    \small
    \centering
    \begin{tabular}{ p{1.80cm}  c c c  } \hline
    \textbf{Dataset} & Question & Answer & Context\\ \hline
    \multicolumn{3}{l}{\textit{Average Length (Tokens)}} \\
    ~~SearchQA & 17.25 & 31.51 & 55.50 \\
    ~~TriviaQA & 15.56 & 33.88 & 747.75 \\
    ~~HotpotQA & 18.52 & 28.31 & 91.57 \\
    ~~SQuAD & 11.45 & 29.70 & 140.64 \\
    ~~NQ & 9.24 & 107.10 & 220.02 \\
    ~~BioASQ & 11.18 & 29.01 & 241.52 \\
    ~~R.E. & 9.15 & 27.51 & 29.14 \\
    ~~TextbookQA & 10.20 & 16.37 & 648.23  \\
    \hline
    \multicolumn{3}{l}{\textit{Question/Answer Token Overlap (\%)}} \\
    ~~SearchQA & - & 37.83 & 55.23 \\
    ~~TriviaQA & - & 25.53 & 74.23 \\
    ~~HotpotQA & - & 29.08 & 49.16 \\
    ~~SQuAD & - & 43.03 & 56.36 \\
    ~~NQ & - & 23.50 & 36.87 \\
    ~~BioASQ & - & 23.08 & 53.40 \\
    ~~R.E. & - & 39.21 & 40.98 \\
    ~~TextbookQA & - & 25.64 & 82.54  \\
    \hline
    \end{tabular}
    \caption{Average length (\# of word tokens) and degree of question/answer token overlap of each constructed dataset.}
    \label{tab:dataset_analysis}
\end{table}

% =====================================================================

\section{Models}
\label{sec:models}

Two neural models, based on BERT~\cite{devlin-etal-2019-bert} and USE-QA~\cite{yang2019multilingual}, respectively, are evaluated on the MultiReQA test suite. Performance is contrasted with a strong term-based information retrieval baseline, BM25.

\subsection{BERT}

\begin{figure}[!htb]
  \centering
  \includegraphics[width=0.48\textwidth]{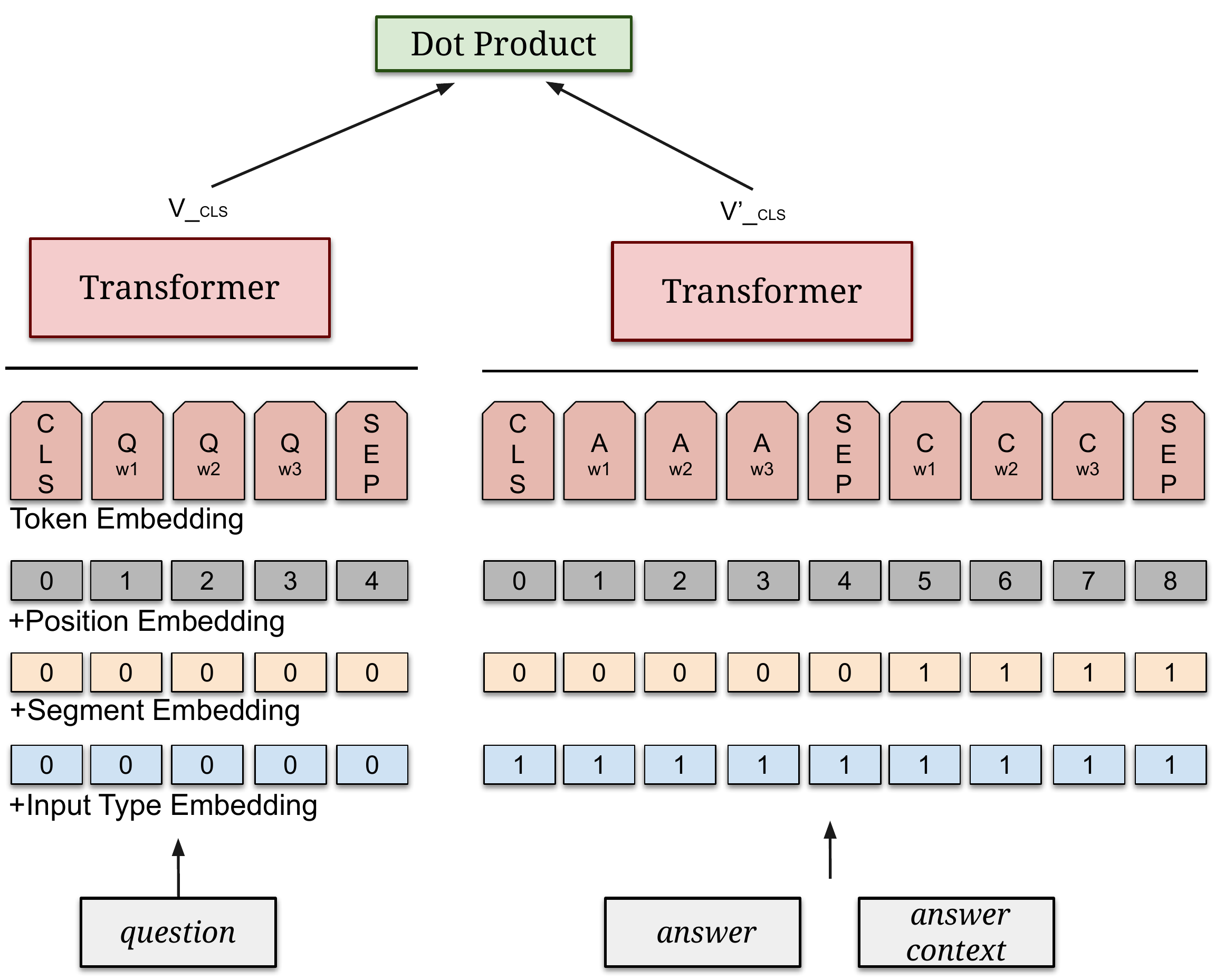}
  \caption{
    The BERT dual encoder architecture. The answer and context are concatenated and fed into the answer encoder. % An additional \textit{input type embedding} is used to differentiate question and answer.
  }
  \label{fig:bert_dual_encoder}

\end{figure}

Given the strong performance of BERT~\cite{devlin-etal-2019-bert} on many language understanding tasks, we explore adapting BERT into a dual encoder as our first neural baseline.
Figure \ref{fig:bert_dual_encoder} illustrates our BERT dual encoder architecture.
The question and answer are encoded separately.
On the left side, the question is fed into a BERT transformer network, and we take the embedding output of the CLS token as the question encoding.
On the right side, the answer text and context are concatenated as a long sequence, using segment IDs to separate them.
The concatenated input is fed into the same BERT transformer network.
As with the question encoder, we take the CLS embedding as the answer encoding.
To distinguish questions and answers, we add an additional \textit{input type embedding} to each input token.
% The input type ID is always 0 for question tokens and always 1 for answer tokens.
Note that we switch the final activation layer of the BERT CLS token from \textit{tanh} to \textit{gelu}.
The final embeddings are $l_2$ normalized.

We employ the BERT\textsubscript{BASE} model,\footnote{The BERT\textsubscript{BASE} model uses 12 transformer layers with 12 attention heads, a hidden size of 768 and a filter size of 3072.
The final embedding size is 768.} due to memory constraints during training.\footnote{We use in-batch negative sampling in the dual encoder training, which requires relatively large batch size. For more details of dual encoder training with negative sampling, see \citet{gillick2018end} and \citet{guo-etal-2018-effective}.}

\subsection{Universal Sentence Encoder QA}

\begin{figure}[!htb]
  \centering
  \includegraphics[width=0.38\textwidth]{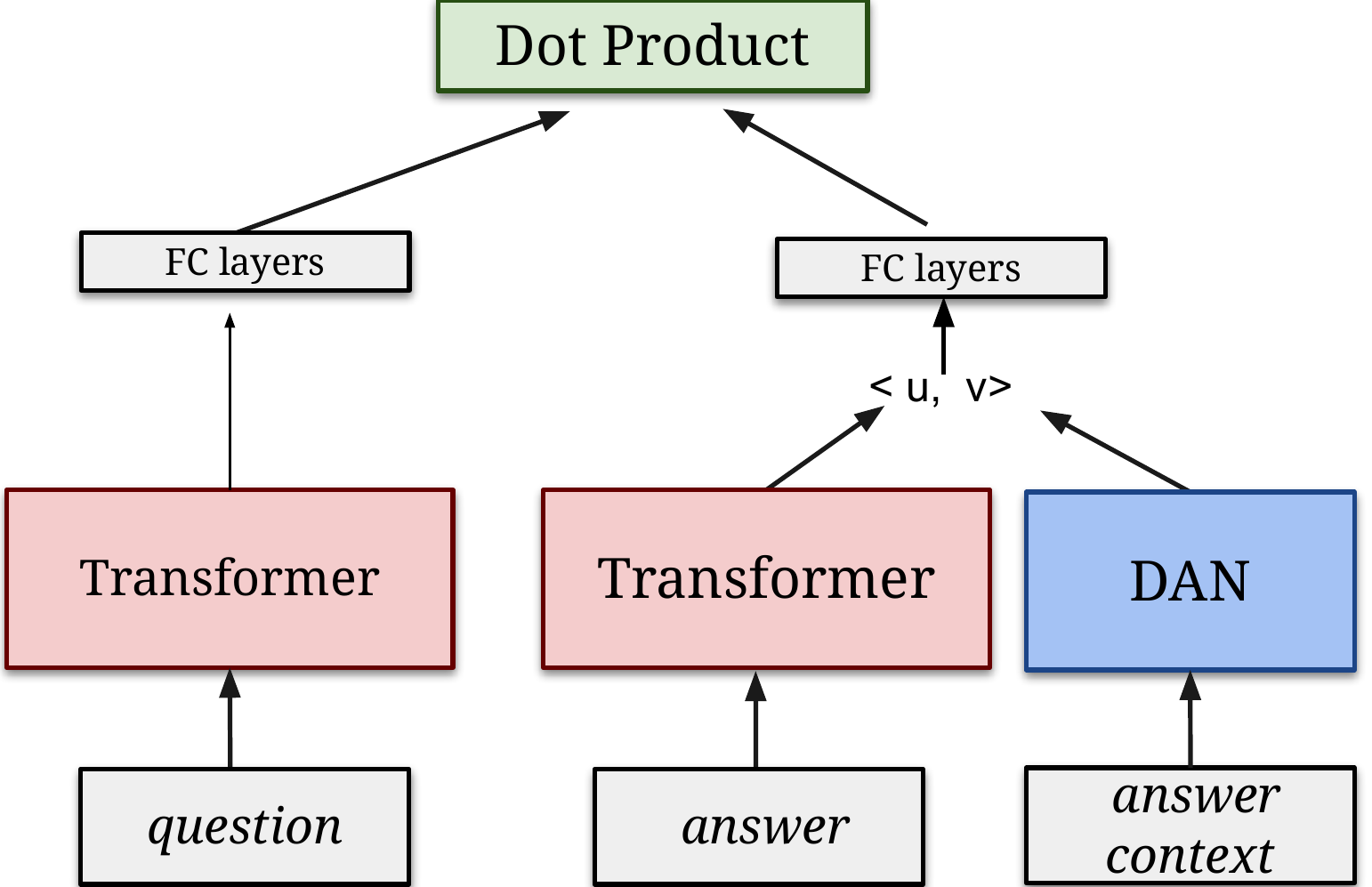}
  \caption{
    The USE-QA model architecture.
  }
  \label{fig:qa_dual_encoder}
\end{figure}

Following \cite{reqa}, we also employ Universal Sentence Encoder QA (USE-QA)~\cite{yang2019multilingual}\footnote{https://tfhub.dev/google/universal-sentence-encoder-multilingual-qa/1} as a neural baseline.
It is a multilingual QA retrieval model pre-trained on billions of examples from web-crawled question answering corpora.
In USE-QA, the question and answer are encoded separately using a dual encoder architecture.
On the left side, the question is encoded using a transformer~\citep{transformer} network with final average pooling.
The pooled output is then fed into a fully-connected network.
On the right side, the answer text and answer context are encoded using a transformer network and a deep averaging network (DAN)~\cite{iyyer-EtAl:2015:ACL-IJCNLP} respectively.
The answer text transformer encoder is shared with question transformer encoder, and employs average pooling.
Then the answer encoding and context are concatenated as a single vector and fed into another fully-connected network.
Both question and answer embeddings are $l_2$ normalized before being fed into the dot product operation.
The model architecture is illustrated in Figure \ref{fig:qa_dual_encoder}.\footnote{
USE-QA uses a 6 layer transformer with 8 attention heads, a hidden size of 512 and a filter size of 2048.
The context DAN encoder uses hidden sizes [320, 320, 512, 512] with residual connections.
The feed-forward networks for question and answer both use hidden sizes [320, 512], so the final dimension of the encodings is 512.}

\subsection{BM25}

Term frequency inverse document frequency (TF-IDF) based methods remain the dominant method for document retrieval, with the ``Best Matching 25'' (BM25) family of ranking functions providing a strong baseline~\citep{Robertson:2009}.
In previous work on open domain question answering, BM25 has been used to retrieve evidence text, and has been shown to be a particularly strong baseline on tasks where the question is written with advance knowledge of the answer~\citep{lee-etal-2019-latent}.

The BM25 score of document~$D$ given query~$Q$ which contains words $q_1, ..., q_n$ is given by:
\begin{equation} 
\label{eq:bm25} 
\small
    % \text{S}(D,Q) =
    \sum_{i=1}^{n} \text{IDF}(q_i) \cdot \frac{f(q_i, D) \cdot (k_1 + 1)}{f(q_i, D) + k_1 \cdot (1 - b + b \cdot \frac{|D|}{\text{avgdl}})} 
\end{equation}
where $f(q_i, D)$ is $q_i$'s term frequency in the document, $|D|$ is the length of the document in words, and $avgdl$ is the average document length across all documents.
Scalars $k_1$ and $b$ are free parameters.

We concatenate the answer sentence and context as the document when applying BM25 for answer retrieval. 
In this setup, the answer sentence is duplicated twice in its context.
Thus, the score for each sentence in context remains unique.

%{\color{red} Decide which version(s) of BM25 to use.}

%--- BM25 SQuAD R@1 ---\\
%Lowercased WPM: 57.3\\
%Lowercased NLTK: 56.0\\
%Original case WPM: 54.6\\
%Original case NLTK: 51.0

% =====================================================================

\section{Experiments}
\label{sec:experiments}

\begin{table*}[htb!]
    \small
    \centering
    \begin{tabular}{ l |l | r r r r r | r r r } \hline
    \multirow{2}{*}{Metric} & \multirow{2}{*}{Models}& \multicolumn{5}{c|}{In-domain Datasets} & \multicolumn{3}{c}{Out-of-domain Datasets} \\ 
    %\cline{3-11}
    \rule{-2pt}{8pt}
    & &  SearchQA & TriviaQA & HotpotQA & NQ & SQuAD & BioASQ & R.E. & TextbookQA \\ \hline
    \multirow{5}{*}{P@1} 
    & BM25\textsubscript{word}                      & 30.94 & 39.35 & 21.04 & 10.07 & 61.50 & 6.38 & 55.75 & 8.39 \\
    & BM25\textsubscript{wpm}                       & \textbf{35.86} & \textbf{43.26} & 20.37 & 25.32 & 65.32 & \textbf{8.31} & \textbf{64.04} & \textbf{8.52} \\ 
    % & DualEncoder                                   & 25.99 & 19.81 & 14.13 & 25.10 & 28.38 & 2.32 &  7.35 & 32.49 & 3.39 \\
    & USE-QA                                        & 31.17 & 28.60 & 18.12 & 24.71 & 51.02 & 5.58 & 52.05 & 7.52 \\
    %& USE-QA\textsubscript{finetune}                & 31.45 & 32.58 & 31.71 & \textbf{38.00} & \textbf{66.83} & - & - & - & - \\
    & USE-QA\textsubscript{finetune}                & 31.45 & 32.58 & 31.71 & \textbf{38.00} & \textbf{66.83} & 6.41 & 59.87 & 6.62 \\
    %& BERT\textsubscript{finetune}                  & 30.20 & 29.11 & \textbf{32.05} & 36.22 & 55.13 & - & - & - & - \\
    & BERT\textsubscript{finetune}                  & 30.20 & 29.11 & \textbf{32.05} & 36.22 & 55.13 & 5.71 & 49.89 & 6.29 \\
    % & BERT\textsubscript{finetune}                  & 28.29 & 28.41 & 34.80 & 39.29 & 57.06 & 5.57 & 15.69 & 50.58 & 6.59 \\  % Finetune on all
    \hline
    \multirow{5}{*}{MRR} 
    & BM25\textsubscript{word}                      & 47.75 & 51.58 & 33.07 & 15.51 & 69.16 & 10.37 & 71.27 & 17.23 \\
    & BM25\textsubscript{wpm}                       & \textbf{52.25} & \textbf{55.80} & 32.99 & 37.1 & 72.96 & 12.86 & \textbf{79.86} & 16.97 \\ 
    % & DualEncoder                                   & 39.79 & 30.66 & 21.79 & 36.66 & 38.24 & 5.69 & 12.92 & 46.35 & 8.84 \\
    & USE-QA                                        & 47.52 & 40.26 & 22.65 & 34.73 & 62.08 & 12.31 & 67.41 & 16.92 \\
    %& USE-QA\textsubscript{finetune}                & 50.70 & 42.39 & 43.77 & \textbf{52.27} & \textbf{75.86} & - & - & - & - \\
    & USE-QA\textsubscript{finetune}                & 50.70 & 42.39 & 43.77 & \textbf{52.27} & \textbf{75.86} & 13.39 & 74.89 & 15.49 \\
    & BERT\textsubscript{finetune}                  & 47.08 & 41.34 & \textbf{46.21} & 52.02 & 64.74 & \textbf{19.21} & 65.21 & \textbf{20.17} \\
    % & BERT\textsubscript{finetune}                  & 47.57 & 46.05 & 49.74 & 56.11 & 69.11 & 19.00 & 36.31 & 67.64 & 22.65 \\ % Finetune on all
    \hline
    \end{tabular}
    \caption{Precision at 1(P@1)(\%) and Mean Reciprocal Rank (MRR)(\%) on the constructed question answer retrieval datasets. USE-QA\textsubscript{finetune} and BERT\textsubscript{finetune} are fine-tuned on each in-domain dataset individually.
    The performance of fine-tuned models on out-of-domain datasets are the average score across all five fine-tuned models.}
    \label{tab:results}
\end{table*}

\subsection{Fine-tuning and Configurations}

We use the BM25 implementation in the Gensim library~\cite{gensim} with default $k_1$ and $b$ settings.
Inverse document frequency is calculated for each constructed dataset independently.
We deploy two different tokenization methods for BM25: NLTK~\cite{nltk} and a WordPiece model (wpm)~\cite{wu2016google} following the BERT implementation\footnote{The wpm vocab is from BERT\textsubscript{BASE}.}.
Note that NLTK does not normalize the text, while the WordPiece model does. 
We also experimented on SQuAD with removing normalization from wpm, and found that wpm still outperforms NLTK\@.
Our results in Table~\ref{tab:results} for BM25\textsubscript{word} use NLTK without normalization, while BM25\textsubscript{wpm} uses wpm with normalization.

The USE-QA model was pre-trained specifically for retrieval question answering tasks.
So we first evaluate the default model without any dataset specific fine-tuning. 
We further fine-tune the USE-QA model using the same discriminitive objective for retrieval used for the original USE-QA training~\cite{yang2019multilingual}:
\begin{equation}
\label{eq:bayes}
P(y \mid x) = \frac{e^{\phi(x, y)}} {\sum_{\bar{y} \in \mathcal{Y}} e^{\phi(x, \bar{y})}}
\end{equation}
Where $x$ is the question, $y$ is the correct answer, $\mathcal{Y}$ is all answers in the same batch that are used as sampled negatives, and $\phi(x, y)$ is the dot product of question and answer representations.

We fine-tune each USE-QA model on the in-domain training set using batch size 64, and SGD optimizer with learning rate decaying exponentially from 0.01 to 0.001.
All model are trained 10 epochs.

BERT was pre-trained for masked language modeling and next-sentence prediction, rather than for retrieval.
To adapt BERT for retrieval, we fine-tune our BERT dual encoder, with the same discriminative objective used to fine-tune the USE-QA models.
We use in-batch random negative sampling with batch size 128, and the default AdamW optimizer with learning rate 0.0001.
Each BERT based model is trained 10 epochs. Note that neural model hyper-parameters are tuned on a validation set (10\%) split out from the training data.

\begin{table*}[htb!]
    \small
    \centering
    \begin{tabular}{ l | l | r r r r r | r r r } \hline
    \multirow{2}{*}{Metric} &
    \multirow{2}{*}{\diagbox[innerwidth=1.8cm]{Train}{Test}} &
    \multicolumn{5}{c|}{In-domain Datasets} & \multicolumn{3}{c}{Out-of-domain Datasets} \\ 
    %\cline{3-11}
    \rule{-2pt}{8pt}
       & & SearchQA & TriviaQA & HotpotQA & NQ & SQuAD & BioASQ & R.E. & TextbookQA \\ \hline
    \multirow{7}{*}{P@1}
    & SearchQA  & 31.45 & 35.48 & 16.04 & 24.69 & 46.60 & 6.52 & 60.03 & 6.66 \\
    & TriviaQA  & 28.44 & 32.58 & 14.91 & 22.58 & 38.87 & 4.45 & 60.84 & 4.06 \\
    & HotpotQA  & 30.79 & 32.70 & \textbf{31.71} & 26.45 & 56.17 & 5.65 & 57.21 & 6.52 \\
    & NQ        & 28.80 & 31.77 & 17.64 & \textbf{38.00} & 52.23 & 6.52 & 55.48 & 7.66 \\
    & SQuAD     & 31.44 & 35.21 & 20.25 & 28.32 & \textbf{66.83} & \textbf{7.65} & \textbf{63.73} & 8.32 \\
    \cdashline{2-10} \rule{-2pt}{8pt}
    & Joint   & \textbf{32.24} & 37.40 & 26.54 & 36.35 & 60.81 & 7.58 & 62.71 & 7.52 \\
    & Joint\textsubscript{No TriviaQA} & 31.92 & \textbf{37.71} & 29.68 & 36.23 & 64.00 & 6.78 & 61.69 & \textbf{8.72} \\
    \hline 
    \hline
    \multirow{7}{*}{MRR}
    & SearchQA  & 50.70 & 47.88 & 25.88 & 36.31 & 57.83 & 13.34 & 75.51 & 15.19\\
    & TriviaQA  & 44.57 & 42.39 & 23.40 & 32.77 & 47.50 & 9.26 & 75.88 & 10.49 \\
    & HotpotQA  & 47.17 & 44.41 & \textbf{43.77} & 36.99 & 66.25 & 32.15 & 72.54 & 15.08 \\
    & NQ        & 45.08 & 44.39 & 26.57 & \textbf{52.27}  & 62.88 & 13.77 & 70.07 & 17.71\\
    & SQuAD     & 48.70 & 48.16 & 30.12 & 38.79 & \textbf{75.86} & \textbf{15.75} & \textbf{78.50} & \textbf{18.71}\\
    \cdashline{2-10} \rule{-2pt}{8pt}
    & Joint   & \textbf{51.04} & \textbf{50.88} & 38.95 & 50.11 & 71.02 & 14.86 & 78.05 & 16.61 \\
    & Joint\textsubscript{No TriviaQA} & 50.80 & 50.77 & 41.62 & 49.93 & 73.71 & 14.69 & 77.04 & 18.64 \\
    \hline
    \end{tabular}
    \caption{P@1(\%) and MRR(\%) of USE-QA models fine-tuned on either one or all in-domain datasets, evaluated across all datasets.  \textbf{Joint}: Fine-tune on all in-domain datasets together. \textbf{Joint\textsubscript{No TriviaQA}}: Same as ``Joint'', but removing TriviaQA from the fine-tuning data pool.}
    \label{tab:results_transfer}
\end{table*}

\subsection{Results}

Table \ref{tab:results} shows baseline model performance of precision at 1 (P@1) and Mean Reciprocal Rank (MRR) on the constructed retrieval QA datasets.
The highest score for each task is bolded.
For P@1, the first two rows shows the results for BM25\textsubscript{word} and BM25\textsubscript{wpm}.
Notably, BM25\textsubscript{wpm} performs better on 7 of 8 tasks, indicating that a careful selection of tokenization and normalization can improve the term-based model considerably.
The advantage of BM25\textsubscript{wpm} is particularly noticeable on datasets where the question is constructed without seeing the answer: SearchQA, TriviaQA, NQ, BioASQ and Relation Extraction.
BM25\textsubscript{wpm} also achieves the highest P@1 on 2 of 5 in-domain datasets and on all out-of-domain dataests.

% The third row shows the results of pre-trained USE-QA model, and rows 4-5 are USE-QA and BERT dual encoder fined-tuned with training data: USE-QA\textsubscript{finetune} and BERT\textsubscript{finetune} respectively. 
The remaining rows show the results of the neural models: the off-the-shelf USE-QA model, as well as fine-tuned versions of USE-QA and the BERT dual encoder model.
We finetune on each in-domain dataset separately, and the performance on the out-of-domain datasets is the average across all five fine-tuned models.
The default USE-QA model is overall not competitive with BM25\textsubscript{wpm}.
However when fine-tuned on in-domain data, USE-QA outperforms BM25\textsubscript{wpm} on 3 of 5 in-domain datasets.
For most datasets, fine-tuned BERT (pre-trained over generic text) performed nearly as well fine-tuned USE-QA model (pre-trained over question-answer pairs).
This indicates that it is not critical to pre-train on question answering data specifically.
However, large-scale pre-training is still critical, as we will see in section \ref{sec:pretraining}.

We observe that the best neural models outperform BM25\textsubscript{wpm} on Hotpot and NQ by large margins: +11.68 and +12.68 on P@1 respectively. 
This result aligns with the statistics from Table~\ref{tab:dataset_stats}, where token overlap between question and answer/context is low for these sets.
For datasets with high overlap between question and answer/context, BM25\textsubscript{wpm} performs better than neural models.
% This observation is also aligned with the hypothesis that BM25 is good at term matching, while neural models have potential to go beyond.

The same conclusion for P@1 can be drawn for MRR, with the exception that BERT\textsubscript{finetune} outperforms the other models on BioASQ and TextbookQA.
We observe that the vocabulary of BioASQ and TextbookQA are different from the other datasets, including more specialized technical terms.
Comparing with other models, the good MRR performance of BERT\textsubscript{finetune} may be due to the better token embedding from the masked language model pre-training.

\subsection{Transfer Learning across Domains}

The previous section shows that neural models are competitive when training on in-domain data, with USE-QA slightly outperforming the BERT dual encoder.
In order to better understand how fine-tuning data helps the neural models, in this section we experiment with training on different datasets, focusing on the USE-QA fine-tuned model. 
Table \ref{tab:results_transfer} shows the performance of models trained on each individual dataset, as well as a model trained jointly on all available in-domain datasets.

Each column compares the performance of different models on a specific test set.
The highest numbers of each test set are bolded. %and the underline numbers are the second highest.
Rows 1 through 5 show the results of the models trained on each in-domain dataset.
In general, models trained on an individual dataset achieve the best (or near-best) performance on their own eval split, with the exception of TriviaQA.
It is interesting to see that the model fine-tuned on TriviaQA performs poorly on nearly all datasets.
This suggests the sentence-level training data quality from TriviaQA might be lower than other datasets.
TriviaQA requires reasoning across multiple sources of evidence~\cite{joshi-etal-2017-triviaqa}, so sentences with annotated answer spans may not directly answer the posed question.

Rows 6 and 7 are models trained on the combined datasets.
In addition to the model trained jointly on all the datasets, we also train a model without TriviaQA, given the poor performance of the model trained individually on this set.
The model trained over all available data is competitive, but the performance on some datasets, e.g.~NQ and SQuAD, is significantly lower than the individually-trained models.
By removing TriviaQA, the combined model gets close to the individual model performance on NQ and SQuAD, and achieves the best P@1 performance on TriviaQA and TextbookQA.

% =====================================================================

\section{Analysis}
\label{sec:analysis}

\subsection{Does Context Help?}

Candidate answers may be not fully interpretable when taken out of their surrounding context~\cite{reqa}.
In this section we investigate how model performance changes when removing context.
We experiment with one BM25 model and one neural model, by picking the best performing models from previous experiments: BM25\textsubscript{wpm} and USE-QA\textsubscript{finetune}.
Recall, USE-QA\textsubscript{finetune} models are fine-tuned on each individual dataset.

\begin{figure}[!htb]
  \centering
  \includegraphics[width=0.48\textwidth]{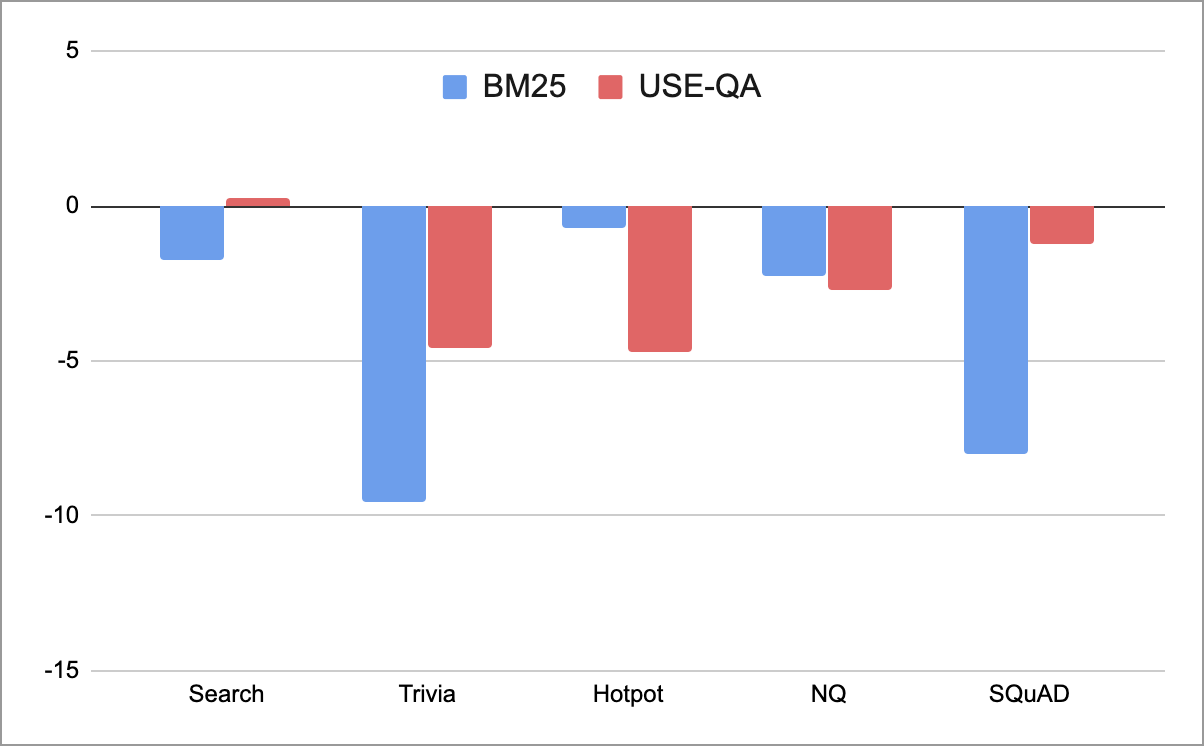}
  \caption{
    Performance change in P@1(\%) of the BM25\textsubscript{wpm} and USE-QA\textsubscript{finetune} models when we remove the surrounding context.
  }
  \label{fig:qa_no_context}
\end{figure}

Figure \ref{fig:qa_no_context} illustrates the change in performance when models are restricted to only use the candidate answer sentence.\footnote{We report P@1 here, but observed similar trends in MRR.}
Even without the surrounding context, both the BM25 model and USE-QA model are still able to retrieve many of the correct answers.
For the USE-QA model, the performance drop is less than 5\% on all datasets. The drop in BM25 performance is larger, supporting the hypothesis that BM25's token overlap heuristic is most effective over large spans of text, while the neural model is able to provide a ``deeper'' semantic understanding and squeeze more signal out of a single sentence.
% [...] that the BM25 model relies more on the tokens, while the neural model is more likely to understand the semantic meaning of the sentence.

\subsection{Pre-training}
\label{sec:pretraining}

\begin{table}[!t]
    \small
    \centering
    \begin{tabular}{l | r | r r } \hline
    \multirow{3}{*}{\textbf{Models}}& \multirow{3}{*}{\makecell{\textbf{No} \\\textbf{Pre-training}}} & \multicolumn{2}{c}{\textbf{USE-QA Pre-training}} \\ % & \multicolumn{2}{c}{ICT pre-training} \\ 
    \rule{-2pt}{8pt}
                & & \makecell{without \\ fine-tuning} & \makecell{with\\ fine-tuning} \\ \hline % & no fine-tuning & with fine-tuning \\ \hline
    Search    & 25.99 & 31.17 & 32.24 \\ % & 25.54 & 28.63 \\
    Trivia    & 19.81 & 28.60 &  37.4 \\ % & 13.24 & 29.3 \\
    Hotpot    & 14.13 & 18.12 & 26.54 \\ % & 13.21 & 25.00 \\
    NQ        & 25.10 & 24.71 & 36.35 \\ % & 13.74 & 31.46 \\
    SQuAD     & 28.38 & 51.02 & 60.81 \\ % & 44.58 & 56.51 \\
    \hline
    BioASQ    &  2.32 &  5.58 & 7.78  \\ % &  4.85 & 5.72 \\
    % DuoRC     &  7.35 & 22.67 & 23.27 \\ % & 14.97 & 19.83 \\
    R.E.      & 32.49 & 52.05 & 62.71 \\ % & 39.76 & 59.35 \\
    Textbook  &  3.39 &  7.52 &  7.52 \\ % &  4.99 & 6.66\\
    \end{tabular}
    \caption{P@1(\%) of models with/without pre-training.}
    \label{tab:pretrain}
\end{table}

We perform a simple ablation by training the USE-QA model architecture directly on the fine-tuning data, starting from randomly initialized parameters. The results are shown in Table \ref{tab:pretrain}.
The ``No Pre-training'' model is trained jointly on all available in-domain data, as we found that training from scratch on individual datasets performed even worse.
The model performs worse than the out-of-the-box pre-trained model on seven of the eight datasets, and is dramatically worse across all datasets compare to the fine-tuned pre-trained model.
This indicates that large-scale pre-training is critical for getting good QA retrieval performance from neural models.
However, recalling the strong performance of the BERT dual encoder in Table~\ref{tab:results}, it is not critical to pre-train over question answering data specifically.

\subsection{Error Analysis}

\begin{table*}[htb!]
\small    \centering
     \begin{tabular}{p{0.97\linewidth}}  \hline \hline
     \rule{-2pt}{12pt}
     % \textbf{NQ} \\ \hline
     \textbf{Example 1} (from NQ): what kind of fish live in the salton sea \\
     \rule{-2pt}{12pt}
     \textbf{Correct Answer}: [...] Due to the high salinity , very few fish species can tolerate living in the Salton Sea . \textit{Introduced tilapia are the main fish that can tolerate the high salinity levels and pollution .} Other freshwater fish species live in the rivers and canals that feed the Salton Sea , including threadfin shad . [...] \\
     \rule{-2pt}{12pt}
     \textbf{USE-QA\textsubscript{finetune}}: [...] It may also drift in to the south - western part of the Baltic Sea ( where it can not breed due to the low salinity ) . \textit{Similar jellyfish -- which may be the same species -- are known to inhabit seas near Australia and New Zealand .} The largest recorded specimen found washed up on the shore of Massachusetts Bay in 1870 . [...] \\
     \rule{-2pt}{12pt}
     \textbf{BM25\textsubscript{wpm}}: [...] Introduced tilapia are the main fish that can tolerate the high salinity levels and pollution . \textit{Other freshwater fish species live in the rivers and canals that feed the Salton Sea , including threadfin shad , carp, red shiner , channel catfish , white catfish , largemouth bass , mosquitofish , sailfin molly , and the vulnerable desert pupfish .} [...] \\[4pt]
     \hline \\[-5pt]
     \rule{-2pt}{4pt}
     \textbf{Example 2} (from TriviaQA): What was invented in the 1940s by Percy Spencer, an American self-taught engineer from Howland, Maine, who was building magnetrons for radar sets? \\
     \rule{-3pt}{12pt}
     \textbf{Correct Answer}: [...] After experimenting, he realized that microwaves would cook foods quickly - even faster than conventional ovens that cook with heat. \textit{The Raytheon Corporation produced the first commercial microwave oven in 1954; it was called the 1161 Radarange.} It was large, expensive, and had a power of 1600 watts. [...] \\
     \rule{-2pt}{12pt}
     \textbf{USE-QA\textsubscript{finetune}}: [...] Because of his accomplishments, Spencer was awarded the Distinguished Service Medal by the U.S. Navy and has a building named after him at Raytheon. \textit{Percy Spencer, while working for the Raytheon Company, discovered a more efficient way to manufacture magnetrons.} In 1941, magnetrons were being produced at a rate of 17 per day. [...] \\
     \rule{-2pt}{12pt}
     \textbf{BM25\textsubscript{wpm}}: [...] By the end of 1971, the price of countertop units began to decrease and their capabilities were expanded. \textit{Spencer, born in Howland, Maine, was orphaned at a young age.} Although he never graduated from grammar school, he became Senior Vice President and a member of the Board of Directors at Raytheon, receiving 150 patents during his career [...]
     \\[5pt]
     \hline\hline % \rule{0pt}{4pt}
 \end{tabular}
     \caption{Examples where both the BM25\textsubscript{wpm} and USE-QA\textsubscript{finetune} models get wrong. \textit{Italics} indicate the answer sentence. At most one sentence before/after the answer is shown, although the original context may be longer.}
     \label{tab:error_analysis}
 \end{table*}

In this section we examine some typical failure cases of the BM25\textsubscript{wpm} and USE-QA\textsubscript{finetune} models.
As a first observation, the two models retrieve very different answers.
For example, we find that on Natural Questions, the two models' top-ranked answers disagree on 64.75\% questions\footnote{Note that even if the models retrieve different answers, both answers could still be correct.}.
The other datasets have similar levels of disagreement.
This suggests that the models have different strengths, and that a combination of these modeling techniques could leads to a significant improvement. % We leave this as future work. 

% and it requires contextual understanding to deduce the connection between introduced tilapia, high salinity levels, and Salton Sea.

Table \ref{tab:error_analysis} shows examples where the models retrieve different answers, and both are incorrect. In the first example, the BM25\textsubscript{wpm} retrieves the correct context by matching the keyword ``Salton Sea''. But it fails to retrieve the correct sentence, as none of the keywords in the question appear in the target answer. On the other hand, the USE-QA\textsubscript{finetune} model understands the question is asking about some sort of animal living in the sea, but fails to connect to the Salton Sea specifically.
Similarly, in the second example, both models retrieve sentences that match some keywords from the question. The BM25\textsubscript{wpm} matches keywords ``Spencer'' and ``Maine'', but misses that the question is looking for an invention. The USE-QA\textsubscript{finetune} matches ``Spencer'', and is able to connect ``invent'' with ``discover'', but surfaces the wrong discovery.

Overall, we observe that the term based model is able to retrieve the correct context in most cases, but often fails to select the correct answer sentence, as that sentence may not have the highest token matching score with the question. On the other hand, the neural model seems to ``understand'' the question a little better, but sometimes fails to recognize important keywords.

\section{Related Work}
\label{sec:related_word}

Open domain QA answers questions by querying a large collection of documents~\citep{trec}.
Existing open domain QA datasets usually measure if a system's output matches the ground-truth answer of the given question, often a word or a short phrase.
For example, the DrQA~\citep{chen-etal-2017-reading} task treats Wikipedia as a knowledge base to answer factoid questions from SQuAD~\citep{rajpurkar-etal-2016-squad}, CuratedTREC~\citep{curatedTREC}, and other datasets.
The task measures how well a system can successfully extract a string containing the answer to a question.
Instead, our work follows ReQA task, and differs from this type of task by retrieving a complete sentence-level answer.

Similar to ReQA task, \citet{seo-etal-2018-phrase} constructs a phrase-indexed QA challenge benchmark retrieving phrases, allowing for a direct $F_1$ and exact-match evaluation on SQuAD\@. 
An extended work~\cite{seo-etal-2019-real} demonstrates the phrase-indexed QA system can be built using a combination of dense (neural) and sparse (term-frequency based) indices. 
\citet{lareqa} investigates the retrieval of sentence-level answers from a language agnostic candidate pool.
\citet{Chang2020Pre-training} investigates the pre-training tasks for retrieving answers from a large scale candidate pool.

Finally, \citet{surdeanu-etal-2008-learning} provides a dataset consisting of 142,627 question-answer pairs from Yahoo!~Answers ``how to'' questions, with the goal of retrieving the correct answer to a given question from the set of all answers.
WikiQA~\citep{yang-etal-2015-wikiqa} is another sentence-level answer selection dataset consisting of 3,047 questions and 29,258 candidate answers, split into train, dev, and test.
These datasets, however, are either limited to a specific type of question, or limited to a small set of candidates.

We propose a more comprehensive eval covering multiple domains and include tasks at a much larger scale.
Additionally, folding the various MRQA in-domain and out-of-domain datasets into a single eval allows us to directly investigate cross-domain generalization.

% =====================================================================

\section{Conclusion}
\label{sec:conclusion}
In this paper, we convert eight existing QA tasks from the MRQA shared task~\cite{fisch2019mrqa} into retrieval versions, by treating the sentence containing the ground-truth span as the target sentence-level answer.
We establish baselines using unsupervised term-based information retrieval methods (the BM25 ranking function), as well as two supervised neural models built on pre-trained USE-QA and BERT models.
Overall, a classical term-based retrieval approach, BM25, is a strong baseline, and could likely be improved further using additional information retrieval techniques such as normalization and synonym handling.
The neural models, however, can be trained end-to-end without much feature engineering, and perform particularly well on tasks with a low degree of question/answer token overlap, or in situations where the context length is limited.
The neural model performance can also be improved through the addition of in-domain training data. However, we find that QA tasks are not all alike and having training data in the precise target domain is important.

\section*{Acknowledgements}
We thank our teammates from Descartes and other Google groups for their feedback and suggestions, particularly DK Choe and Kelvin Guu.

\bibliography{acl2020}
\bibliographystyle{acl_natbib}

\end{document}